\documentclass[
]{ceurart}

\usepackage{listings}
\begin{document}

\copyrightyear{2025}
\copyrightclause{Copyright for this paper by its authors.
  Use permitted under Creative Commons License Attribution 4.0
  International (CC BY 4.0).}

\conference{CLEF 2025 Working Notes, 9 -- 12 September 2025, Madrid, Spain}

\title{CEA-LIST at CheckThat! 2025: Evaluating LLMs as Detectors of Bias and Opinion in Text}

\title[mode=sub]{Notebook for the CheckThat! Lab at CLEF 2025}


\author[1]{Akram Elbouanani}[
orcid=0009-0005-0597-7149,
email=elbouanani.akram@gmail.com
]
\address[1]{Université Paris-Saclay, CEA-List, F-91120, Palaiseau, France}

\author[1]{Evan Dufraisse}[
email=evan.dufraisse@cea.fr
]
\author[1]{Aboubacar Tuo}[
email=aboubacar.tuo@cea.fr
]

\author[1]{Adrian Popescu}[
email=adrian.popescu@cea.fr
]

\begin{abstract}
This paper presents a competitive approach to multilingual subjectivity detection using large language models (LLMs) with few-shot prompting. We participated in Task 1: Subjectivity of the CheckThat! 2025 evaluation campaign. We show that LLMs, when paired with carefully designed prompts, can match or outperform fine-tuned smaller language models (SLMs), particularly in noisy or low-quality data settings. Despite experimenting with advanced prompt engineering techniques—such as debating LLMs and various example selection strategies—we found limited benefit beyond well-crafted standard few-shot prompts. Our system achieved top rankings across multiple languages in the CheckThat! 2025 subjectivity detection task, including first place in Arabic and Polish, and top-four finishes in Italian, English, German, and multilingual tracks. Notably, our method proved especially robust on the Arabic dataset, likely due to its resilience to annotation inconsistencies. These findings highlight the effectiveness and adaptability of LLM-based few-shot learning for multilingual sentiment tasks, offering a strong alternative to traditional fine-tuning, particularly when labeled data is scarce or inconsistent.

\end{abstract}


\maketitle

\section{Introduction}

In natural language processing, distinguishing between subjective and objective language is a foundational task with wide-ranging implications. Subjectivity refers to expressions that convey personal opinions, beliefs, sentiments, or evaluations, whereas objectivity is characterized by verifiable facts and observable phenomena independent of individual interpretation \cite{ruggeri2023definition}. Accurate subjectivity detection plays a crucial role in applications such as fact-checking, media analysis, and content moderation, where separating opinion from factual reporting is essential to counter misinformation and uphold media integrity \cite{yu2003towards}. In legal domains, detecting subjective language in witness testimonies or contractual texts can inform interpretation and decision-making \cite{make6020041}. Similarly, in sentiment analysis, identifying subjective expressions enables a deeper understanding of public attitudes toward products, services, or events \cite{make6020041, liu2022sentiment}.

The emergence of Large Language Models (LLMs) has presented new opportunities for text analysis tasks. Unlike traditional Smaller Language Models (SLMs), which often require extensive, task-specific fine-tuning on large annotated datasets, LLMs leverage their pre-training to address numerous tasks with minimal, or even no, additional training \cite{brown2020language}. Their ability to recognize subtle linguistic cues, such as irony or sarcasm, makes them particularly suited for identifying subjective expressions. Furthermore, their adaptability enables them to conform more precisely to specific annotation guidelines, offering flexibility in defining and identifying subjective content.

In previous editions of the CheckThat! lab, the application of LLMs for tasks such as subjectivity detection has been less prevalent compared to fine-tuned SLMs \cite{galassi2023overview, struss2024overview}. In this context, in the rare instances where they were used, their performance has not consistently exceeded that of optimized SLMs \cite{paran2024semanticcuetsync}. This work aims to demonstrate that LLMs can indeed compete with, and potentially outperform, fine-tuned SLMs in subjectivity detection tasks through the application of techniques such as careful prompting and few-shot prompting. We investigate the extent to which precise prompt engineering and the provision of relevant examples within the prompt can enhance the performance of LLMs, thereby showcasing their potential for robust and adaptable subjectivity detection in academic and real-world settings. While these techniques show promise in theory, our experiments reveal that they do not consistently improve performance on the tested datasets, suggesting important limitations and directions for further research.

\section{Related Work}

The computational detection of subjective language has evolved significantly from its early rule-based foundations to contemporary data-driven approaches. This progression reflects broader trends in natural language processing while addressing challenges in identifying opinionated content. Current research emphasizes two critical aspects: (1) the development of increasingly sophisticated models capable of capturing linguistic nuance \cite{javdan2020applying, zhang-etal-2024-sentiment}, and (2) the creation of specialized techniques to optimize these models for subjectivity analysis \cite{shokri2024subjectivity, huang2025structuredreasoningfairnessmultiagent, suwaileh2024thatiar}.

\textbf{Architectural Evolution.} Traditional lexicon-based and supervised learning approaches have given way to transformer-based models, with BERT-style architectures demonstrating strong performance on binary subjectivity classification \cite{8834477, 9197910, kotelnikova2021lexicon}. However, the emergence of LLMs has introduced new capabilities in detecting implicit subjectivity through contextual reasoning. Indeed, a key advantage of LLMs is their advanced ability to recognize subtle linguistic cues like irony, sarcasm, or implicit bias \cite{potamias2020transformer}. Their architectural design enables them to capture complex dependencies within sequential data, leading to a deeper understanding of intricate relationships between words and emotions. Recent empirical studies have demonstrated that LLMs consistently achieve higher overall accuracy in sentiment analysis, often outperforming specialized pre-trained transformer models due to their comprehensive grasp of human thought and emotion \cite{zhang-etal-2024-sentiment}. This ability indicates strong potential for subjectivity detection, as recognizing nuanced evaluative language is key to distinguishing subjective from objective statements \cite{ruggeri2023definition}.

\textbf{Prompting Strategies.} Despite their inherent capabilities, optimizing LLM performance in specialized tasks like subjectivity detection often requires appropriate prompting. While a wide range of prompting strategies exists, we focus in this work on a selected subset of strategies which we evaluate in our experiments.

\begin{itemize}
    \item \textbf{Prompt engineering} involves meticulously designing input queries to guide LLMs toward desired outputs. This technique is crucial for clearly defining the nuances of subjectivity for the model, specifying output formats, and aligning the LLM's reasoning with task-specific objectives \cite{marvin2023prompt, sahoo2024systematic}. However, prompt effectiveness can be highly sensitive to wording, and small changes may lead to inconsistent results \cite{zhuo2024prosaassessingunderstandingprompt, errica2025didiwrongquantifying}.
    
    \item \textbf{In-context Learning (ICL)} refers to the ability of LLMs to perform tasks by conditioning on input-output examples provided directly in the prompt, without updating model parameters. A common subcategory of ICL is \textbf{few-shot learning}, where the prompt includes a small number of labeled examples to help the model infer the task and classification criteria \cite{wang2020generalizing, brown2020language}. A key challenge in few-shot ICL is selecting representative and diverse examples, as LLMs can overfit to or ignore suboptimal demonstrations.
    
    \item \textbf{Multi-agent LLM Systems} is an emerging paradigm to enhance LLM performance. This approach distributes responsibilities across multiple specialized agents, each focusing on specific functions like information retrieval, complex reasoning, or decision-making \cite{du2024improving}. Multi-agent systems offer several advantages, including enhanced reliability through cross-verification, refined decision-making through collaborative information sharing, and improved handling of complex tasks by dividing workloads \cite{du2024improving, liang-etal-2024-encouraging, shinn2023reflexion}. Yet, coordination overhead and potential inconsistencies between agents remain significant challenges \cite{du2024improving}.

\end{itemize}

The CheckThat! lab \cite{CheckThat:ECIR2025, clef-checkthat:2025-lncs}, organized within CLEF, serves as a prominent platform for advancing subjectivity detection research, particularly in distinguishing subjective from objective statements at the sentence level within news articles across multiples languages \cite{struss2024overview}. CheckThat! evaluations have systematically demonstrated the strengths and limitations of different approaches to fact-checking and subjectivity detection. In early iterations (2018-2020), traditional machine learning models, such as SVM with carefully engineered linguistic features, achieved competitive results, particularly for English texts \cite{atanasova2018overview, atanasova2019overview}. The 2020-2021 evaluations marked a transition period in which fine-tuned BERT-style models began to dominate the leaderboards \cite{shaar2020overview, shaar2021overview}. These results established transformer architectures as the new baseline for subjectivity detection tasks.
The most recent CheckThat! cycles (2023-2024) included early approaches relying on LLMs. While early submissions underperformed due to inadequate prompting strategies, subsequent systems demonstrated that properly optimized LLMs could match or exceed specialized models \cite{galassi2023overview, struss2024overview}.

\section{Dataset}

We utilize the provided multilingual dataset, which comprises sentence-level annotations labeled as either \textsc{OBJ} (objective) or \textsc{SUBJ} (subjective). Table~\ref{tab:dataset-stats} summarizes the number of annotated sentences per language and split. The dataset exhibits class imbalance across languages and splits, with some languages (e.g., Italian and Arabic) showing a predominance of \textsc{OBJ} labels, while others (e.g., Bulgarian) present a more balanced distribution.

During the exploratory phase, we focus primarily on English and Arabic. These two languages were selected due to their differing class distributions and dataset sizes. We operate under the assumption that insights obtained from these languages are transferable to the other languages in the dataset.

\begin{table}[ht]
\centering
\caption{Number of sentences and class distribution (\textsc{OBJ} vs \textsc{SUBJ}) across languages and data splits.}
\begin{tabular}{lcccc}
\toprule
\textbf{Language} & \textbf{Split} & \textbf{Sentences} & \textbf{OBJ} & \textbf{SUBJ} \\
\midrule
\multirow{4}{*}{English} 
    & Train     & 830  & 532  & 298  \\
    & Dev       & 462  & 222  & 240  \\
    & Dev-Test  & 484  & 362  & 122  \\
\midrule
\multirow{4}{*}{Italian} 
    & Train     & 1613 & 1231 & 382  \\
    & Dev       & 667  & 490  & 177  \\
    & Dev-Test  & 513  & 377  & 136  \\
\midrule
\multirow{4}{*}{German} 
    & Train     & 800  & 492  & 308  \\
    & Dev       & 491  & 317  & 174  \\
    & Dev-Test  & 337  & 226  & 111  \\
\midrule
\multirow{4}{*}{Bulgarian} 
    & Train     & 729  & 406  & 323  \\
    & Dev       & 467  & 175  & 139  \\
    & Dev-Test  & 250  & 143  & 107  \\
\midrule
\multirow{4}{*}{Arabic} 
    & Train     & 2446 & 1391 & 1055 \\
    & Dev       & 742  & 266  & 201  \\
    & Dev-Test  & 748  & 425  & 323  \\
\bottomrule
\end{tabular}
\label{tab:dataset-stats}
\end{table}

\section{Methodology}

We explore several strategies to improve subjective sentence classification. We experiment with the following three main approaches: prompt engineering, few-shot learning, and multi-agent LLM setups. We also use fine-tuned SLMs as baselines.

\subsection{Prompt Engineering}

We systematically evaluate the impact of prompt phrasing and label framing on classification performance. We compare:
\begin{itemize}
    \item A minimal, generic prompt vs. a detailed one generated from the annotation guidelines.
    \item Label framing using explicit terms (``Subjective''/``Objective'') vs. neutral terms (``Category 0''/``Category 1'').
    \item Binary yes/no questions (e.g., ``Is the sentence subjective?'') as an alternative to direct classification.
\end{itemize}
These variations are designed to probe how linguistic framing affects the model's interpretability and consistency.

\subsection{Few-Shot Learning Strategies}

We experiment with a range of few-shot prompting configurations to assess how the number and selection of support examples influence performance. Specifically, we compare:

\begin{itemize}
\item Prompting setups with 0 (no-shot), 6-shot, and 12-shot examples.
\item Example selection strategies based on:
(a) Semantic similarity,
(b) Semantic dissimilarity, and
(c) Random sampling.
\end{itemize}

For the semantic approaches, similarity and dissimilarity are measured using cosine similarity between sentence embeddings generated by OpenAI’s \textit{text-embedding-3-small} model. To ensure class balance in the few-shot prompts, we selected an equal number of examples from each class (subjective and objective). For instance, in the 6-shot setting, the prompt included the three most similar (or dissimilar) subjective examples and the three most similar (or dissimilar) objective ones. This balance helps prevent prompt-induced bias toward a particular class label.

\subsection{Multi-Agent LLM Reasoning}

We design a set of multi-agent prompting experiments to investigate the interpretability and robustness of LLM outputs:
\begin{itemize}
    \item \textbf{Debate setup:} Two agents argue why a sentence is subjective vs. objective; a third model acts as a judge.
    \item \textbf{Adversarial reasoning:} One agent argues why the sentence is not subjective and another why it is not objective, and a judge makes the final call.
    \item \textbf{Extended framing:} We include all four perspectives---Subjective, Not Subjective, Objective, and Not Objective--- and a judge making the final decision.
\end{itemize}

\section{Results}

This section presents a detailed evaluation of multiple modeling strategies for subjectivity classification, including fine-tuned transformers, prompted LLMs with and without few-shot examples, advanced prompt reframing, and agent-based debating approaches. Given the dataset's imbalance, the official primary evaluation metric is the macro-averaged F1-score, which equally weights both classes. We also pay particular attention to \textsc{SUBJ} recall, as the subjective class is often underrepresented and may be more informative in downstream analyses. This imbalance was mitigated using a weighted BCE loss, warmup training, and early stopping to ensure stability and generalizability. We first report results with different system variants on the development subset and then discuss the official results obtained with the 2025 test set.

\subsection{Preliminary Results}

\subsubsection{Fine-Tuned Transformers}

Table~\ref{tab:finetuned} summarizes the performance of supervised transformer models trained on English and Arabic data. RoBERTa-Base, fine-tuned on English data, achieves the best performance overall, with a macro F1 of 0.70 and a notably high macro precision of 0.79. However, the model struggles with subjective instances, achieving only 0.39 recall for the subjective class, which suggests a strong bias toward the majority (objective) class.

In Arabic, the results are considerably weaker across the board. While BERT-Base-Arabertv02 achieves the best Arabic macro F1 score (0.55), subjective recall remains modest (0.47). Despite the use of language-specific models and XLM-RoBERTa for cross-lingual encoding, the performance gap between English and Arabic remains substantial.

\begin{table}[h]
\centering
\caption{Fine-tuned transformer performance on subjectivity classification in English and Arabic.}
\begin{tabular}{lcccccc}
\toprule
\textbf{Model} & \textbf{Setup} & \textbf{Lang} & \textbf{Macro F1} & \textbf{Macro P} & \textbf{P Subj} & \textbf{R Subj} \\
\midrule
RoBERTa-Base & 10e, 5e-6 lr, 32 bs & English & \textbf{0.70} & 0.79 & 0.76 & 0.39 \\
XLM-RoBERTa & 10e, 5e-6 lr, 32 bs & English & 0.66 & 0.74 & 0.68 & 0.34 \\
BERT-Base-Arabertv02 & 10e, 5e-5 lr, 16 bs & Arabic & \textbf{0.55} & 0.55 & 0.50 & 0.47 \\
XLM-RoBERTa & 10e, 5e-6 lr, 16 bs & Arabic & 0.51 & 0.54 & 0.51 & 0.25 \\
\bottomrule
\end{tabular}
\label{tab:finetuned}
\end{table}

\subsubsection{Prompt Engineering and Few-Shot Learning}

Table~\ref{tab:prompting} presents results from zero- and few-shot prompting using GPT-4o-mini. A basic prompt yields poor subjective precision (0.32) but strong recall (0.67), suggesting that under-specification of the task leads the model to over-predict subjectivity. By contrast, a more detailed prompt derived from annotation guidelines improves both macro F1 (+0.12) and subjective precision (+0.14).

Incorporating six or twelve few-shot examples drawn randomly further boosts performance, reaching a macro F1 of 0.76. The improvement is consistent across both subjective precision and recall. Using 12-shot few-shot learning with extended prompting slightly improves recall but saturates macro F1 gains.

\begin{table}[h]
\centering
\caption{Prompt-based and few-shot GPT-4o-mini performance in English.}
\begin{tabular}{lcccc}
\toprule
\textbf{System} & \textbf{Macro F1} & \textbf{Macro P} & \textbf{P Subj} & \textbf{R Subj} \\
\midrule
GPT-4o-mini (Basic Prompt) & 0.54 & 0.57 & 0.32 & 0.67 \\
GPT-4o-mini (Extended Prompt) & 0.66 & 0.65 & 0.46 & 0.56 \\
+ FSL (6-shot, Random) & \textbf{0.76} & 0.78 & 0.69 & 0.60 \\
+ FSL (12-shot, Random) & \textbf{0.76} & 0.77 & 0.66 & 0.63 \\
\bottomrule
\end{tabular}
\label{tab:prompting}
\end{table}

\subsubsection{Few-Shot Selection Strategies}

We further compare several strategies for selecting few-shot examples: random sampling, similarity-based sampling, and dissimilarity-based sampling. In similarity-based sampling, we choose examples that are most similar to the test sentence, while in dissimilarity-based sampling, we select those that are the most different. Similarity is measured using the cosine similarity between sentence embeddings generated by GPT-3. Results are shown in Table~\ref{tab:fslselection}.

Interestingly, random selection outperforms similarity-based strategies across all models. For GPT-4o-mini, random sampling yields the best macro F1 (0.76), whereas dissimilarity-based selection offers a better recall (0.73) but slightly lower overall F1. A similar trend is observed for Qwen-72B, where dissimilar sampling boosts recall (+0.07 over similarity) but offers minimal F1 gain. This suggests that dissimilar examples may help capture broader linguistic variance, aiding generalization. These results contrast with earlier findings highlighting the benefits of semantically similar examplars for in-context learning \cite{xu2023knnpromptingbeyondcontextlearning}.

\begin{table}[h]
\centering
\caption{Few-shot selection strategies across LLMs in English.}
\begin{tabular}{lcccc}
\toprule
\textbf{System} & \textbf{Macro F1} & \textbf{Macro P} & \textbf{P Subj} & \textbf{R Subj} \\
\midrule
\textbf{GPT-4o-mini} & & & & \\
\quad + Random & \textbf{0.76} & 0.78 & 0.69 & 0.60 \\
\quad + Similarity & 0.70 & 0.69 & 0.52 & 0.62 \\
\quad + Dissimilarity & 0.75 & 0.74 & 0.57 & \textbf{0.73} \\
\midrule
\textbf{LLaMA 70B} & & & & \\
\quad + Random & 0.73 & 0.73 & 0.61 & 0.57 \\
\quad + Similarity & 0.70 & 0.71 & 0.58 & 0.51 \\
\quad + Dissimilarity & \textbf{0.75} & 0.77 & 0.67 & 0.31 \\
\midrule
\textbf{Qwen 72B} & & & & \\
\quad + Random & 0.71 & 0.71 & 0.55 & 0.60 \\
\quad + Similarity & 0.71 & 0.70 & 0.52 & 0.67 \\
\quad + Dissimilarity & \textbf{0.73} & 0.72 & 0.57 & 0.64 \\
\bottomrule
\end{tabular}
\label{tab:fslselection}
\end{table}

\subsubsection{Prompt Reframing and Debate-Based Inference}

We investigate whether the way labels are framed affects model behavior. Reframing ``subjective vs. objective'' as a binary question (e.g., ``Is the sentence subjective? Yes/No'') or as category labels (``Category 1 vs. Category 2'') leads to slight F1 gains over the base prompt (Table~\ref{tab:reframing}). Framing clearly influences the model’s inductive bias, with category labels yielding better subjective precision (0.69) and macro F1 (0.72).

Debating-based prompting (Table~\ref{tab:debating}) also provides strong results. The setup where one LLM argues for subjectivity, another for objectivity, and a judge decides, achieves the best macro F1 overall (0.77). Notably, this format significantly enhances subjective recall (up to 0.74), suggesting that reasoning-focused prompting facilitates more balanced decisions. Debate variants using negated prompts (e.g., ``Not Subjective'' vs. ``Not Objective'') also perform competitively.

\begin{table}[h]
\centering
\caption{Effect of prompt reframing with GPT-4o-mini in English.}
\begin{tabular}{lcccc}
\toprule
\textbf{Framing Strategy} & \textbf{Macro F1} & \textbf{Macro P} & \textbf{P Subj} & \textbf{R Subj} \\
\midrule
Yes/No Binary & 0.71 & 0.70 & 0.52 & 0.70 \\
Category 1 vs 2 & \textbf{0.72} & 0.76 & 0.69 & 0.47 \\
\bottomrule
\end{tabular}
\label{tab:reframing}
\end{table}

\begin{table}[h]
\centering
\caption{Performance of debating-style LLMs with GPT-4o-mini in English.}
\begin{tabular}{lcccc}
\toprule
\textbf{Debating Setup} & \textbf{Macro F1} & \textbf{Macro P} & \textbf{P Subj} & \textbf{R Subj} \\
\midrule
Subjective vs Objective & \textbf{0.77} & 0.76 & 0.62 & 0.72 \\
Not Subjective vs Not Objective & 0.76 & 0.75 & 0.59 & \textbf{0.74} \\
Full Scale (Pos/NPos/Neg/NNeg) & 0.74 & 0.73 & 0.56 & 0.74 \\
\bottomrule
\end{tabular}
\label{tab:debating}
\end{table}

\subsubsection{LLM Ensemble Results}

Finally, we evaluate an ensemble voting strategy that aggregates predictions from five diverse models: RoBERTa-Base, GPT-4o-mini, LLaMA 70B, Qwen 72B, and Aya-Expanse 32B. As shown in Table~\ref{tab:ensemble}, this ensemble achieves the highest overall macro F1 score (0.79), with a strong subjective precision of 0.77. These results indicate that ensembling models with heterogeneous architectures and training paradigms can effectively capture complementary perspectives on subjectivity, enhancing robustness and performance.

\begin{table}[h]
\centering
\caption{Ensemble classification performance in English.}
\begin{tabular}{lcccc}
\toprule
\textbf{System} & \textbf{Macro F1} & \textbf{Macro P} & \textbf{P Subj} & \textbf{R Subj} \\
\midrule
LLM Ensemble & \textbf{0.79} & 0.77 & 0.77 & 0.59 \\
\bottomrule
\end{tabular}
\label{tab:ensemble}
\end{table}

\subsection{Final Results}

\subsubsection{Evaluation Setup}

The official evaluation of the CheckThat! 2025 campaign comprised three settings:

\begin{itemize}
    \item \textbf{Monolingual}: train and test on data in a given language \(L\) (Arabic, Italian, German, English).
    \item \textbf{Multilingual}: train and test on data comprising several languages.
    \item \textbf{Zero-shot}: train on several languages and test on unseen languages (Romanian, Polish, Ukrainian, Greek).
\end{itemize}

For our final submitted system in the official campaign evaluation, we adopted the extended-prompt strategy using randomly selected 6-shot examples, paired with an ensemble of multiple models, including GPT-4 variants (GPT-4o-mini, GPT-4.1-mini), RoBERTa, LLaMA 70B, and Qwen 72B. In the zero-shot setting, the in-context examples were provided in English, following the task guidelines.

Final results are reported  in Table~\ref{tab:final-results}. 

\begin{table}[h!]
\centering
\caption{Final Macro F1 results for each language including the top-ranked team, \textbf{CEA-LIST} (our team), the baseline, and the last-ranked team. When \textbf{CEA-LIST} is first, the second-ranked team is also reported.}
\begin{tabular}{l l c c}
\hline
\textbf{Language} & \textbf{Team} & \textbf{Rank} & \textbf{Macro F1} \\
\hline
\textbf{Italian} & XplaiNLP & 1 & 0.8104 \\
                 & \textbf{CEA-LIST}    & \textbf{2} & \textbf{0.8075} \\
                 & \textit{Baseline }            & 11 & 0.6941 \\
                 & IIIT Surat            & 14 & 0.4612 \\
\hline
\textbf{Arabic}  & \textbf{CEA-LIST}    & \textbf{1} & \textbf{0.6884} \\
                 & UmuTeam       & 2 & 0.5903 \\
                 & \textit{Baseline}            & 8 & 0.5133 \\
                 & JU\_NLP              & 14 & 0.4328 \\
\hline
\textbf{German}  & smollab              & 1 & 0.8520 \\
                 & \textbf{CEA-LIST}    & \textbf{4} & \textbf{0.7733} \\
                 & \textit{Baseline}             & 15 & 0.6960 \\
                 & IIIT Surat            & 16 & 0.6342 \\
\hline
\textbf{English} & msmadi               & 1 & 0.8052 \\
                 & \textbf{CEA-LIST}    & \textbf{3} & \textbf{0.7739} \\
                 & UGPLN             & 22 & 0.5531 \\
                 &\textit{ Baseline   }          & 23 & 0.5370 \\
\hline
\textbf{Multilingual} & TIFIN India & 1 & 0.7550 \\
                      & \textbf{CEA-LIST}   & \textbf{3} & \textbf{0.7396} \\
                      &\textit{ Baseline }           & 13 & 0.6390 \\
                      & AI Wizards        & 16 & 0.2380 \\
\hline
\textbf{Polish}  & \textbf{CEA-LIST}    & \textbf{1} & \textbf{0.6922} \\
                 & IIIT Surat            & 2 & 0.6676 \\
                 & \textit{Baseline }            & 9 & 0.5719 \\
                 & TIFIN INDIA & 14 & 0.3811 \\
\hline
\textbf{Ukrainian} & CSECU-Learners         & 1 & 0.6424 \\
                   & \textit{Baseline}            & 5 & 0.6296 \\
                   & \textbf{CEA-LIST}   & \textbf{10} & \textbf{0.6061} \\
                   & TIFIN INDIA & 14 & 0.4731 \\
\hline
\textbf{Romanian} & msmadi               & 1 & 0.8126 \\
                  & \textbf{CEA-LIST}    & \textbf{6} & \textbf{0.7659} \\
                  & \textit{Baseline  }           & 13 & 0.6461 \\
                  & TIFIN INDIA & 14 & 0.5181 \\
\hline
\textbf{Greek}    & AI Wizards         & 1 & 0.5067 \\
                  & \textbf{CEA-LIST}    & \textbf{7} &\textbf{ 0.4492} \\
                  & \textit{Baseline}             & 9 & 0.4159 \\
                  & TIFIN India & 14 & 0.3337 \\
\hline
\end{tabular}
\label{tab:final-results}
\end{table}

\subsubsection{Discussion}

Our team demonstrated strong results across multiple languages, achieving first place in Arabic and Polish, and top-three positions in the majority of the evaluated languages, including Italian, English, and multilingual settings. This consistent performance underscores the robustness and generalizability of our approach using LLM-based few-shot learning.

Our experiments demonstrate that leveraging large language models (LLMs) instead of fine-tuned smaller language models (SLMs) can yield highly competitive results across multiple languages and settings. However, the effectiveness of LLMs critically depends on the quality of prompt design. For instance, advanced prompting strategies such as debating LLMs, where multiple model outputs are cross-examined, did not lead to substantial improvements over standard few-shot prompting. Similarly, varying the example selection method, whether by similarity, dissimilarity, or random choice, showed no significant impact on final performance. These findings suggest that while prompt engineering remains essential, more complex example selection or ensemble strategies may not always provide additional gains.

The most notable result was in Arabic, where we outperformed the second-ranked team by a substantial margin of +0.10 Macro F1-score. We attribute this advantage partly to the nature of the Arabic dataset, which exhibits annotation inconsistencies. Unlike fine-tuned models that heavily depend on high-quality labeled training data, our few-shot LLM approach is less affected by such noise. Prior research has indicated that in-context learning with LLMs can be relatively independent of the exact label quality provided in training examples \cite{min-etal-2022-rethinking}. Consequently, our method was more resilient to inconsistencies, resulting in superior evaluation performance. This highlights a significant practical benefit of using LLMs: they can better handle noisy or imperfect datasets, offering an edge in real-world scenarios where high-quality annotations are difficult to obtain.

Overall, our findings suggest that LLMs, combined with carefully crafted few-shot prompts, offer a powerful and flexible alternative to traditional fine-tuning approaches, especially when training data quality varies. This has important implications for future multilingual sentiment analysis tasks and other NLP challenges where data quality and multilingual coverage are key concerns.

\section{Dataset quality}

During our evaluation of the Arabic dataset, we consistently observed limited performance across all tested configurations. Regardless of the prompting strategy employed, ranging from simple to extended prompts, in both zero-shot and few-shot settings, the macro F1-score remained below \textbf{0.55}. This plateau in performance was observed across multiple models, including GPT-4 and fine-tuned BERT models as shown in Table~\ref{tab:finetuned}, and suggests potential issues beyond model capacity or prompt design.

Interestingly, this contrasts with the results reported in the original dataset paper \cite{suwaileh2024thatiar}, where a five-shot setup using a Maximal Marginal Relevance (MMR)-based example selection achieved an F1-score of \textbf{0.80}. In our comparable few-shot setting with GPT-4 and similarity-based example selection, the F1-score reached only \textbf{0.547}, indicating a significant gap in reproducibility.

Upon closer inspection, we identified potential sources of annotation inconsistency. Manual review revealed several instances where the assigned labels did not seem to align with the guidelines outlined in the original paper. For example, the sentence:

\begin{figure}[ht!]
    \centering
    \includegraphics[width=\linewidth]{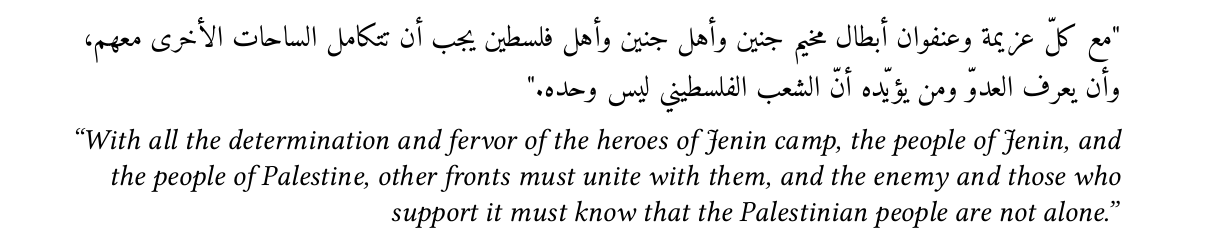}
\end{figure}



is labeled as objective, despite the presence of emotive language and the term "the enemy"
, which could reasonably be interpreted as subjective under the dataset’s own criteria, where they state that such politically charged language must be labeled as subjective. Conversely, clearly factual sentences such as:

\begin{figure}[ht!]
    \centering
    \includegraphics[width=\linewidth]{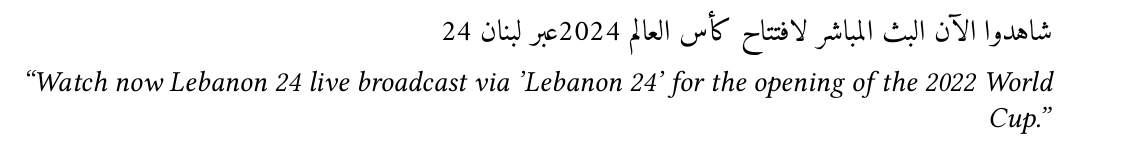}
\end{figure}



are labeled as subjective, despite appearing to report straightforward event announcements. Similarly, sentences comprising purely reported speech, such as the following about COVID-19 statistics, are labeled as subjective even though the annotation guidelines specify otherwise:

\begin{figure}[ht!]
    \centering
    \includegraphics[width=\linewidth]{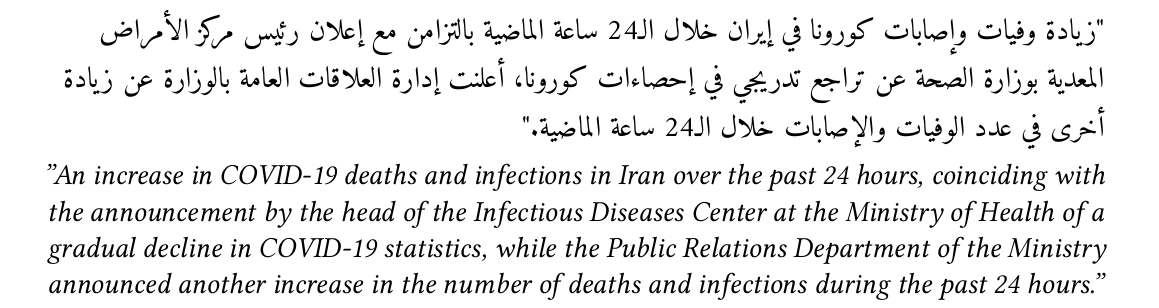}
\end{figure}



To ensure this was not a limitation inherent to the task or language, we ran comparable experiments on the Arabic dataset from the 2023 edition of the task. In that case, our models achieved significantly better performance (F1 = 0.84) using a six-shot extended prompt, and the top team of that year had reported an F1-score of 0.79 \cite{dey2023nn}, demonstrating the feasibility of high performance on well-annotated Arabic datasets.

To further test the hypothesis that label quality rather than linguistic features was the bottleneck, we translated the dataset into English using DeepL and reran the experiments. However, this also did not lead to improved performance (F1 < 0.6), reinforcing our initial hypothesis. A small-scale manual reannotation conducted by one of the authors, who is a native Arabic speaker, led to a moderate increase in performance (F1 = 0.65), providing further evidence that inconsistencies in labeling may play a role in the observed results.

These observations highlight the challenges of subjectivity annotation, especially in politically sensitive contexts, and underline the importance of annotation consistency for benchmarking tasks involving subtle linguistic distinctions.

\section{Conclusion}

In this study, we have shown that large language models (LLMs) used with well-designed few-shot prompting can rival or surpass fine-tuned smaller models (SLMs) across diverse languages and settings. Our approach proved particularly robust in the face of noisy or inconsistent training data, as demonstrated by our strong performance on the Arabic dataset. By consistently ranking among the top teams, securing first place in Arabic and Polish and top-three finishes in most other languages, we illustrate the versatility and effectiveness of LLMs for multilingual subjectivity detection.

While advanced prompt engineering strategies such as debating and varied example selection did not yield major improvements, our results emphasize the critical role of prompt quality itself. The flexibility of LLMs combined with minimal reliance on extensive labeled data offers a promising path forward for multilingual NLP tasks, especially when dealing with data of varying quality.

\begin{acknowledgments}
 This work was supported by the BOOM ANR Project (ANR-20-CE23-0024) and benefited from the use of the FactoryIA supercomputer, funded by the Île-de-France Regional Council. 
\end{acknowledgments}

\section*{Declaration on Generative AI}
This work was assisted by generative AI tools used to improve clarity and style, specifically GPT-4. The authors reviewed and verified all content to ensure accuracy and maintain the integrity of the scientific work.

\bibliography{mybib}

\appendix

\section{Prompts Used}

In this section, we report the prompts used for the classification of the sentences. We report both the simple prompt we experimented with at first and the extended prompt that provided the best performance. We also report the prompts used for the debating LLMs. In particular, we report the prompt used for : (1) the LLM tasked with explaining why a sentence is objective ,(2) the LLM tasked with explaining why a sentence is subjective ,(3) the LLM tasked with explaining why a sentence is not objective, (4) the LLM tasked with explaining why a sentence is not subjective and (5) the judge LLM.

\subsection{Simple Prompt (English):}

You are a linguistic expert, able to detect whether a sentence is objective (OBJ) or subjective (SUBJ). Answer only with \texttt{OBJ} or \texttt{SUBJ}.

\subsection{Extended Prompt (English):}

You are a linguistic expert specializing in detecting whether a sentence is objective or subjective. Your task is to classify sentences according to the following criteria:

\begin{itemize}
    \item \textbf{Objective:} A sentence is objective if it presents factual information, even if the information is debatable or controversial. Additionally:
    \begin{itemize}
        \item \textbf{Emotions:} Statements conveying emotions should be labeled as objective if they reflect the author's beliefs or sensations that cannot be fact-checked or rephrased in a more neutral form.
        \item \textbf{Quotes:} If a sentence contains a direct quote, label it as objective, since the task concerns only the subjectivity of the article’s author, not the quoted speaker. I repeat: \textbf{SENTENCES WHICH ONLY CONTAIN REPORTED SPEECH SHOULD NEVER BE LABELED SUBJECTIVE.}
    \end{itemize}
    \item \textbf{Subjective:} A sentence is subjective if it reflects personal opinions, interpretations, or evaluations. Indicators of subjectivity include:
    \begin{itemize}
        \item \textbf{Intensifiers:} Words or phrases that amplify a statement (e.g., ‘so damaged’) can indicate subjectivity, as they may reflect the author's personal perspective.
        \item \textbf{Speculations:} Statements that imply uncertainty, predictions, or unverifiable claims should be labeled as subjective. For example, phrases like ‘will hope to sow uncertainty’ suggest an interpretation rather than a fact.
    \end{itemize}
\end{itemize}

Answer only with the words \texttt{objective} or \texttt{subjective} based on these criteria.

\noindent
\textbf{Note:} For other languages, this extended prompt was translated using DeepL to ensure semantic accuracy and consistency.

\subsection{Subjectivity Explanation Prompt:}

You are a linguistic expert specializing in detecting whether a sentence is objective (OBJ) or subjective (SUBJ). Your task is to classify sentences according to the following criteria:

\begin{itemize}
    \item \textbf{Objective:} A sentence is objective if it presents factual information, even if the information is debatable or controversial. Additionally:
    \begin{itemize}
        \item \textbf{Emotions:} Statements conveying emotions should be labeled as objective if they reflect the author's beliefs or sensations that cannot be fact-checked or rephrased in a more neutral form.
        \item \textbf{Quotes:} If a sentence contains a direct quote, label it as objective, since the task concerns only the subjectivity of the article’s author, not the quoted speaker. I repeat: \textbf{SENTENCES WHICH ONLY CONTAIN REPORTED SPEECH SHOULD NEVER BE LABELED SUBJECTIVE.}
    \end{itemize}
    
    \item \textbf{Subjective:} A sentence is subjective if it reflects personal opinions, interpretations, or evaluations. Indicators of subjectivity include:
    \begin{itemize}
        \item \textbf{Intensifiers:} Words or phrases that amplify a statement (e.g., ``so damaged'') can indicate subjectivity, as they may reflect the author's personal perspective.
        \item \textbf{Speculations:} Statements that imply uncertainty, predictions, or unverifiable claims should be labeled as subjective. For example, phrases like ``will hope to sow uncertainty'' suggest an interpretation rather than a fact.
    \end{itemize}
\end{itemize}

Given the following sentence, explain why it is classified as \textbf{subjective} based on these criteria. Try to be concise and explain why it is classified as such. Do not repeat the sentence in your answer. Keep to the annotation guidelines given above. Maintain a critical mindset, you can disagree with the classification, but do so only if you are certain. Do not speculate about the sentence’s intention.

\subsection{Objectivity Explanation Prompt:}

You are a linguistic expert specializing in detecting whether a sentence is objective (OBJ) or subjective (SUBJ). Your task is to classify sentences according to the following criteria:

\begin{itemize}
    \item \textbf{Objective:} A sentence is objective if it presents factual information, even if the information is debatable or controversial. Additionally:
    \begin{itemize}
        \item \textbf{Emotions:} Statements conveying emotions should be labeled as objective if they reflect the author's beliefs or sensations that cannot be fact-checked or rephrased in a more neutral form.
        \item \textbf{Quotes:} If a sentence contains a direct quote, label it as objective, since the task concerns only the subjectivity of the article’s author, not the quoted speaker. I repeat: \textbf{SENTENCES WHICH ONLY CONTAIN REPORTED SPEECH SHOULD NEVER BE LABELED SUBJECTIVE.}
    \end{itemize}
    
    \item \textbf{Subjective:} A sentence is subjective if it reflects personal opinions, interpretations, or evaluations. Indicators of subjectivity include:
    \begin{itemize}
        \item \textbf{Intensifiers:} Words or phrases that amplify a statement (e.g., ``so damaged'') can indicate subjectivity, as they may reflect the author's personal perspective.
        \item \textbf{Speculations:} Statements that imply uncertainty, predictions, or unverifiable claims should be labeled as subjective. For example, phrases like ``will hope to sow uncertainty'' suggest an interpretation rather than a fact.
    \end{itemize}
\end{itemize}

Given the following sentence, explain why it is classified as \textbf{objective} based on these criteria. Try to be concise and explain why it is classified as such. Do not repeat the sentence in your answer. Keep to the annotation guidelines given above. Maintain a critical mindset, you can disagree with the classification, but do so only if you are certain. Do not speculate about the sentence’s intention.

\subsection{Non-Subjectivity Explanation Prompt:}

You are a linguistic expert specializing in detecting whether a sentence is objective (OBJ) or subjective (SUBJ). Your task is to classify sentences according to the following criteria:

\begin{itemize}
    \item \textbf{Objective:} A sentence is objective if it presents factual information, even if the information is debatable or controversial. Additionally:
    \begin{itemize}
        \item \textbf{Emotions:} Statements conveying emotions should be labeled as objective if they reflect the author's beliefs or sensations that cannot be fact-checked or rephrased in a more neutral form.
        \item \textbf{Quotes:} If a sentence contains a direct quote, label it as objective, since the task concerns only the subjectivity of the article’s author, not the quoted speaker. \textbf{SENTENCES WHICH ONLY CONTAIN REPORTED SPEECH SHOULD NEVER BE LABELED SUBJECTIVE.}
    \end{itemize}
    
    \item \textbf{Subjective:} A sentence is subjective if it reflects personal opinions, interpretations, or evaluations. Indicators of subjectivity include:
    \begin{itemize}
        \item \textbf{Intensifiers:} Words or phrases that amplify a statement (e.g., ``so damaged'') can indicate subjectivity, as they may reflect the author's personal perspective.
        \item \textbf{Speculations:} Statements that imply uncertainty, predictions, or unverifiable claims should be labeled as subjective. For example, phrases like ``will hope to sow uncertainty'' suggest an interpretation rather than a fact.
    \end{itemize}
\end{itemize}

Given the following sentence, explain why it should \textbf{not} be classified as \textbf{subjective} based on these criteria. Try to be concise and explain why it does not fit the criteria for subjectivity. Do not repeat the sentence in your answer. Focus only on why it fails to meet the conditions for subjectivity.

\subsection{Non-Objectivity Explanation Prompt:}

You are a linguistic expert specializing in detecting whether a sentence is objective (OBJ) or subjective (SUBJ). Your task is to classify sentences according to the following criteria:

\begin{itemize}
    \item \textbf{Objective:} A sentence is objective if it presents factual information, even if the information is debatable or controversial. Additionally:
    \begin{itemize}
        \item \textbf{Emotions:} Statements conveying emotions should be labeled as objective if they reflect the author's beliefs or sensations that cannot be fact-checked or rephrased in a more neutral form.
        \item \textbf{Quotes:} If a sentence contains a direct quote, label it as objective, since the task concerns only the subjectivity of the article’s author, not the quoted speaker. \textbf{SENTENCES WHICH ONLY CONTAIN REPORTED SPEECH SHOULD NEVER BE LABELED SUBJECTIVE.}
    \end{itemize}
    
    \item \textbf{Subjective:} A sentence is subjective if it reflects personal opinions, interpretations, or evaluations. Indicators of subjectivity include:
    \begin{itemize}
        \item \textbf{Intensifiers:} Words or phrases that amplify a statement (e.g., ``so damaged'') can indicate subjectivity, as they may reflect the author's personal perspective.
        \item \textbf{Speculations:} Statements that imply uncertainty, predictions, or unverifiable claims should be labeled as subjective. For example, phrases like ``will hope to sow uncertainty'' suggest an interpretation rather than a fact.
    \end{itemize}
\end{itemize}

Given the following sentence, explain why it should \textbf{not} be classified as \textbf{objective} based on these criteria. Try to be concise and explain why it does not fit the criteria for subjectivity. Do not repeat the sentence in your answer. Focus only on why it fails to meet the conditions for objectivity.

\subsection{Judge Prompt:}

You are a judge LLM tasked with determining whether a sentence is objective (OBJ) or subjective (SUBJ) based on opinions defending different points of view. Your job is to evaluate these opinions according to the following criteria:

\begin{itemize}
    \item \textbf{Objective (OBJ):} A sentence is objective if it presents factual information, even if the information is debatable or controversial. Additionally:
    \begin{itemize}
        \item \textbf{Emotions:} Statements conveying emotions should be labeled as objective if they reflect the author's beliefs or sensations that cannot be fact-checked or rephrased in a more neutral form.
        \item \textbf{Quotes:} If a sentence contains a direct quote, label it as objective, since the task concerns only the subjectivity of the article’s author, not the quoted speaker.
    \end{itemize}

    \item \textbf{Subjective (SUBJ):} A sentence is subjective if it reflects personal opinions, interpretations, or evaluations. Indicators of subjectivity include:
    \begin{itemize}
        \item \textbf{Intensifiers:} Words or phrases that amplify a statement (e.g., \emph{so damaged}) can indicate subjectivity, as they may reflect the author's personal perspective.
        \item \textbf{Speculations:} Statements that imply uncertainty, predictions, or unverifiable claims should be labeled as subjective. For example, phrases like \emph{will hope to sow uncertainty} suggest an interpretation rather than a fact.
    \end{itemize}

    \item \textbf{Edge Cases:}
    \begin{itemize}
        \item \textbf{Emotions:} Although statements carrying emotions convey a subjective point of view, they cannot be verified or confuted by a fact-checking system and are therefore labeled as objective.
        \item \textbf{Quotes:} When authors use quotes to support their thesis, the quoted content may be subjective, but for classification, it should be labeled as objective, focusing only on the article's author.
        \item \textbf{Intensifiers:} The presence of intensifiers can indicate subjectivity, but it's important to assess whether they genuinely reflect the author's perspective or serve a descriptive purpose.
        \item \textbf{Speculations:} Speculative statements should be regarded as subjective, as they often reflect the author's interpretation and not just factual content.
    \end{itemize}
\end{itemize}

Given the sentence and the opinions, your task is to make a final decision and answer only with \texttt{objective} or \texttt{subjective}.

\end{document}